# What's the Price of Monotonicity? A Multi-Dataset Benchmark of Monotone-Constrained Gradient Boosting for Credit PD

*By* PETR KOKLEV

*Financial institutions face a trade-off between predictive accuracy and interpretability when deploying machine learning models for credit risk. Monotonicity constraints align model behavior with domain knowledge, but their performance cost – the "price of monotonicity" – is not well quantified. This paper benchmarks monotone-constrained versus unconstrained gradient boosting for credit probability of default across five public datasets and three libraries. We define the Price of Monotonicity (PoM) as the relative change in standard performance metrics when moving from unconstrained to constrained models, estimated via paired comparisons with bootstrap uncertainty. PoM in AUC ranges from essentially zero to about 2.9%: constraints are almost costless on large datasets (typically <0.2%, often indistinguishable from zero) and most costly on smaller datasets with extensive constraint coverage (around 2–3%). Thus, appropriately specified monotonicity constraints can often deliver interpretability with small accuracy losses, particularly in large-scale credit portfolios.*

* Independent researcher (PhD in Economics). This version: December 2025 (email: koklevp@gmail.com).

# I. Introduction

Credit risk modeling sits at the intersection of statistical methodology, regulatory compliance, and financial decision-making. Financial institutions worldwide rely on probability of default (PD) models to assess borrower creditworthiness, allocate capital, and meet regulatory requirements under frameworks such as the Basel II/III Internal Ratings-Based (IRB) approach. These models must balance predictive accuracy, regulatory interpretability, and fairness.

The rise of machine learning, particularly gradient boosting methods such as XGBoost, LightGBM, and CatBoost, has transformed credit risk modeling by delivering superior discrimination performance compared to traditional logistic regression and linear models. However, this performance gain comes with a fundamental tension: complex machine learning models can violate basic economic intuition, creating challenges for validation, regulatory approval, and stakeholder trust.

Monotonicity constraints offer a principled solution to this tension. By requiring that model predictions respect domain knowledge – ensuring, for example, that higher debt-to-income ratios monotonically increase default risk, or that higher FICO scores monotonically decrease it – constraints align model behavior with economic reasoning while preserving the flexibility of gradient boosting. This alignment facilitates model explainability and enhances model governance.

Despite the intuitive appeal of monotonicity constraints, a critical empirical question remains poorly understood: what is the performance cost of imposing these constraints? In other words, what is the "price of monotonicity" – the reduction in discrimination or calibration metrics that practitioners must accept when moving from unconstrained to constrained models? This question is central to model developers and risk managers who must make informed trade-offs between interpretability and predictive performance. Yet the existing literature offers little systematic evidence on the performance trade-off of monotonicity constraints in gradient boosting across diverse datasets, libraries, and constraint specifications.

This paper fills that gap with a multi-dataset benchmark of monotone-constrained gradient boosting for credit PD. We quantify this trade-off using a Price of Monotonicity (PoM) metric and a paired evaluation design. Across five public credit datasets and three popular boosting libraries, we study how PoM varies with dataset size, default rate, and constraint coverage, and we include a wrong-sign diagnostic to show what happens when constraints are mis-specified. Our results show that constraints are often nearly costless in large portfolios but can be more expensive when many features are constrained in moderate-sized, high-dimensional settings.

The remainder of the paper proceeds as follows: Section II reviews related work; Section III describes the experimental design; Section IV reports results; Section V discusses implications; Section VI concludes.

## II. Related Work

The tension between model interpretability and predictive performance has emerged as a central concern in credit risk modeling, particularly as financial institutions increasingly deploy complex machine learning algorithms for probability of default (PD) estimation. The literature on interpretable machine learning has grown substantially, driven by both regulatory requirements and the practical need for model governance. Rudin (2019) argues forcefully for inherently interpretable models over post-hoc explanations, emphasizing that black-box models with explanations can be misleading and that for high-stakes decisions, we should “stop explaining black box machine learning models” and instead use interpretable models directly. In credit risk contexts, where regulatory scrutiny is intense and model decisions directly affect capital requirements, this perspective has particular resonance. The European Union's GDPR, together with EBA guidelines and national banking regulations, imposes transparency and information obligations for certain automated credit decisions – requiring firms to provide "meaningful information about the logic involved" and to justify model-based decisions to customers and supervisors (Regulation (EU) 2016/679 General Data Protection Regulation 2016). In practice, these requirements create strong incentives for more interpretable modeling approaches.

Several studies have examined the trade-off between predictive performance and model interpretability. Doshi-Velez and Kim (2017) provide a framework for thinking about interpretability in machine learning and argue that interpretability needs are application and human-dependent. Subsequent work has made this more explicit by distinguishing interpretability needs across different stakeholders – such as model developers, end-users, and regulators (Tomsett et al. 2018; Preece et al. 2018; Suresh et al. 2021). In the context of credit risk, regulatory interpretability – the ability to justify model behavior to supervisors – often requires constraints that align model outputs with domain knowledge and economic intuition. Monotonicity constraints represent one class of interpretability-enhancing constraints that have received particular attention in credit risk applications (Chen and Ye 2023; Alonso-Robisco et al. 2025). The economic logic underlying credit risk – that higher debt burdens, more delinquencies, and lower income should monotonically increase default probability – provides a natural domain for such constraints. However, the empirical question of how much predictive performance must be sacrificed to enforce these constraints has received limited systematic investigation across diverse datasets and model implementations.

XGBoost, introduced by Chen and Guestrin (2016), provides support for feature-level monotonicity constraints via modified split-finding and leaf-updating routines. Similar functionality is available in LightGBM (Ke et al. 2017) and CatBoost (Prokhorenkova et al. 2018). Although all three frameworks implement monotonicity by restricting candidate splits and constraining leaf predictions, their algorithmic details differ: XGBoost and LightGBM enforce monotonicity primarily through local restrictions on candidate splits and bounds on leaf values, whereas CatBoost couples symmetric (oblivious) trees with a monotone-specific model-shrinkage mechanism that adjusts leaf values at each boosting iteration to preserve global monotonicity (as described in the respective software documentation and developer notes).

Despite widespread availability of these implementations in popular machine learning libraries, empirical evidence on the performance cost of monotonicity constraints remains sparse and fragmented. Most existing studies focus on single datasets or limited model comparisons, making it difficult to draw generalizable conclusions about when constraints impose negligible versus substantial performance penalties. The literature contains anecdotal evidence that constraints can sometimes improve performance by reducing overfitting (particularly in small-sample settings), but systematic quantification of the trade-off across diverse credit risk contexts is lacking.

The theoretical literature on constrained learning provides some guidance on the fundamental trade-offs. Altendorf et al. (2012) show how prior knowledge about qualitative monotonicities can be encoded as inequality constraints in Bayesian network classifiers, explicitly reducing the effective hypothesis space and improving accuracy in sparse-data regimes. However, the empirical performance implications of these constraints in high-dimensional credit risk settings with varying sample sizes, class distributions, and feature characteristics remain underexplored.

Credit risk modeling has a long tradition in both economics and finance, with models serving dual purposes: accurate probability estimation for risk management and regulatory compliance for capital adequacy calculations. The Basel Accords (Basel Committee on Banking Supervision 2006) establish regulatory capital requirements that depend on PD estimates, creating strong incentives for accurate probability estimation. However, regulatory requirements extend beyond accuracy: models must be validated, documented, and explainable to supervisors. The increasing use of machine learning in credit risk has raised questions about how to balance these requirements. The literature on credit scoring has evolved from traditional statistical models (e.g., logistic regression, linear discriminant analysis) to more flexible machine learning approaches. Lessmann et al. (2015) conduct a large-scale benchmark of 41 classification algorithms on eight retail credit-scoring datasets and find that ensemble methods – particularly heterogeneous ensembles and random forests – achieve superior discrimination relative to traditional techniques. Their study largely abstracts from interpretability and formal regulatory constraints, focusing instead on statistical accuracy and cost-based measures of scorecard performance. Recent work has begun to bridge the gap between machine learning performance and regulatory requirements. Several regulatory reports and academic studies discuss the challenges of using machine learning in credit risk under prudential regulation, emphasizing model interpretability, validation, and governance (e.g. Board of Governors of the Federal Reserve System 2011; European Banking Authority 2023; Bussmann et al. 2021). Similar concerns about explanations and trust arise in other high-stakes decision-support settings such as healthcare (Bussone et al. 2015). However, these contributions typically do not quantify the performance cost of interpretability-enhancing constraints, leaving practitioners without clear guidance on the trade-offs involved.

Lipton (2018) critiques the "mythos of model interpretability" emphasizing that interpretability is a multi-faceted, non-binary property and that different stakeholders have different interpretability needs. In credit risk contexts, monotonicity constraints are one way to move along this spectrum – more interpretable than unconstrained black-box models but potentially less flexible than fully unconstrained models. Empirical studies quantifying this interpretability – performance trade-off remain relatively limited and are often domain-specific (e.g., Marcinkevičs and Vogt 2023; Allen et al. 2024). Rudin (2019), together with the survey in Rudin et al. (2022), reviews several cases in which inherently interpretable models (including constrained additive and rule-based models) match or exceed the accuracy of black-box alternatives when domain knowledge is incorporated effectively. However, this literature does not focus specifically on credit risk or systematically vary constraint coverage across datasets with different characteristics.

Several gaps in the existing literature motivate our study. First, despite the widespread availability of monotonicity constraints in gradient boosting libraries (XGBoost, LightGBM, CatBoost), there is no systematic multi-dataset benchmark quantifying their performance cost in credit risk settings. Existing credit-risk studies that impose monotonicity or other structural constraints on ML models (e.g. Chen and Ye 2023; Alonso-Robisco et al. 2025) typically rely on one or two benchmark datasets and a narrow range of model specifications, which makes it hard to identify when such constraints impose negligible versus substantial performance penalties. Second, there is no widely used protocol for quantifying the "price" of monotonicity that jointly considers discrimination (e.g. AUC, PR-AUC) and calibration (e.g. Brier score, LogLoss). Studies usually report a limited set of metrics, often focusing on discrimination alone, even though accurate PD estimation is central for capital and pricing decisions. Third, prior work does not systematically compare monotonicity constraint implementations across major gradient boosting libraries, despite known algorithmic differences that can affect empirical behavior. This leaves practitioners with little guidance on which implementation to use. Fourth, beyond general domain knowledge principles (e.g. higher delinquency should imply higher PD), the literature offers limited empirical guidance on how to specify monotonic constraints in credit risk models, and the performance cost of misspecified constraints (for example, constraining a feature in the wrong direction) has, to our knowledge, not been systematically quantified in credit risk settings, despite the availability of formal monotonicity tests.

Our study addresses these gaps with a systematic multi-dataset benchmark that:

(i) Compares unconstrained and monotone-constrained gradient boosting across five diverse credit risk datasets spanning different scales, class distributions, and domains (consumer credit, corporate bankruptcy, peer-to-peer lending).

(ii) Employs three popular gradient boosting libraries (XGBoost, LightGBM, CatBoost) to assess implementation-specific effects.

(iii) Uses a paired evaluation design with identical experimental conditions (train/test splits, preprocessing, feature sets, hyperparameter search spaces) to isolate constraint effects.

(iv) Quantifies uncertainty using paired bootstrap resampling to account for correlation between model predictions.

(v) Introduces the "Price of Monotonicity" (PoM) metric as a standardized, relative measure of constraint cost across discrimination and calibration dimensions.

This design provides evidence on key questions: when monotonicity is essentially "free", when it imposes measurable costs, and how those costs vary across libraries, datasets, and performance metrics.

Beyond quantification, we provide practitioner guidance on when constraints are likely to help versus hurt, how to select which features to constrain based on economic reasoning and train-only evidence, and how to interpret constraint effects across different performance metrics. We also include a diagnostic analysis using intentionally mis-specified constraints to quantify the cost of constraint misspecification. This guidance is especially relevant for credit risk modelers operating under regulatory requirements that demand both accuracy and interpretability, helping them make more informed decisions about when and how to incorporate monotonicity constraints into their modeling workflows.

## III. Methods

### *A. Overview and Experimental Design*

We estimate the performance cost of monotonicity constraints using a paired comparison design. For each of the five datasets and three libraries, we fit unconstrained and monotone-constrained models under identical train/test splits, preprocessing, feature sets, and hyperparameter grids.

Replication code and full experiment configuration files are available in the accompanying GitHub repository (Koklev 2025).

### *B. Datasets*

We evaluate monotonicity constraints across five publicly available credit risk datasets that span diverse contexts, scales, and class distributions. This selection enables us to assess whether the price of monotonicity varies systematically with dataset characteristics such as sample size, default rate, and domain (consumer credit, corporate bankruptcy, peer-to-peer lending). All datasets are widely used in credit risk modeling literature, ensuring comparability with prior work and facilitating reproducibility.

TABLE I – DATASET SUMMARY

| Dataset | Description | Train Size | Test Size | Features | Default Rate (%) | Train/Test Split | Temporal Split |
|---|---|---|---|---|---|---|---|
| Taiwan Credit | Credit card default prediction (UCI) | 21,000 | 9,000 | 23 | 22.1 | Stratified 70/30 | No |

| Polish Bankruptcy | Polish companies bankruptcy 3rd year (UCI) | 7,352 | 3,151 | 64 | 4.7 | Stratified 70/30 | No |
|---|---|---|---|---|---|---|---|
| German Credit (Statlog) | Statlog German Credit risk (UCI) | 700 | 300 | 20 | 30.0 | Stratified 70/30 | No |
| Give Me Some Credit | Financial distress within 2 years (Kaggle/OpenML) | 105,000 | 45,000 | 10 | 6.7 | Stratified 70/30 | No |
| Lending Club (Zenodo) | Loan default vs fully paid (Zenodo curated) | 829,347 | 225,277 | 8 | 20.0 | Chronological: train 2007–2015, test 2017–2018 | Yes |

*Notes:* Dataset sources: Taiwan Credit — (Yeh and Lien 2009); Polish Bankruptcy — (Zięba et al. 2016); German Credit — (Hofmann 1994); Give Me Some Credit — (Credit Fusion and Cukierski 2011), Lending Club — (Ariza-Garzón et al. 2024).

Table 1 provides a summary of dataset characteristics. The five datasets range from 1,000 to over 1 million observations, with default rates spanning from 4.7% to 30.0%, capturing both balanced and highly imbalanced scenarios. Feature counts vary from 8 to 64, reflecting different levels of information richness.

The **Taiwan Credit** dataset (Yeh and Lien 2009) contains credit card accounts with features capturing payment history, bill amounts, and demographics. The six months of payment status indicators and corresponding bill and payment amounts make it well-suited for evaluating monotonicity constraints on temporal payment behavior.

The **Polish Bankruptcy** dataset (Zięba et al. 2016) comprises company financial statements with a rich set of financial ratio features, predicting bankruptcy within three years. It is a highly imbalanced corporate credit risk task and provides a good testbed for specifying monotonic constraints across profitability, leverage, and efficiency measures.

The **German Credit (Statlog)** dataset (Hofmann 1994) is a classic small-benchmark consumer credit dataset with both numeric and categorical predictors. It has been extensively used in credit risk work, making our results directly comparable to prior studies.

The **Give Me Some Credit** dataset (Credit Fusion and Cukierski 2011) contains borrower records with features predicting financial distress (90+ days past due) within two years. It represents a moderately imbalanced consumer credit setting, with variables emphasizing payment history and debt burden typical of standard credit scoring models.

The **Lending Club** dataset (Ariza-Garzón et al. 2024) is a large peer-to-peer lending dataset with application-time variables such as income, debt-to-income ratio, FICO score, and loan amount. Its temporal structure requires chronological train/test splits to avoid look-ahead bias and makes it useful for studying monotonicity in a realistic marketplace lending setting.

This diverse collection enables us to examine whether monotonicity constraints impose different costs across dataset scales, imbalance levels, and domains. The mix of consumer credit, corporate bankruptcy, and peer-to-peer lending contexts also allows us to assess generalizability of findings across credit risk applications.

### *C. Data Preprocessing and Train/Test Splits*

We apply minimal preprocessing across all datasets to preserve the raw signal and ensure that performance differences reflect the impact of monotonicity constraints rather than preprocessing artifacts. For each dataset, we standardize the target variable to binary encoding (0 = no default, 1 = default) and handle missing values when present using library-native methods (gradient boosting libraries handle missing values internally). We perform no feature engineering beyond what is necessary for library compatibility—categorical features are passed directly to libraries that support native categorical handling (XGBoost, LightGBM, CatBoost), preserving their original encoding.

Train/test splits are created using a fixed global random seed (42) to ensure reproducibility and, critically, to guarantee that identical splits are used across all model regimes (unconstrained, constrained, and wrong-sign). This paired design eliminates split-induced variance as a confound when comparing model performance. For non-temporal datasets (Taiwan Credit, Polish Bankruptcy,

German Credit, Give Me Some Credit), we use stratified random splits with a 70/30 train/test ratio, preserving class balance in both partitions. For the temporal Lending Club dataset, we employ a chronological split to avoid lookahead bias: training on loans issued from 2007 through 2015, and testing on loans issued in 2017–2018. Loans issued in 2016 are excluded from the analysis to maintain a clean separation between the training and test periods. This temporal split simulates real-world deployment conditions where models are evaluated on future data.

Split details for all datasets, including exact train and test sizes, are reported in Table 1. The identical split structure across all model regimes ensures that any observed performance differences between unconstrained and constrained models can be attributed to the monotonicity constraints themselves, not to differences in data partitioning or preprocessing.

### *D. Monotonicity Constraint Specification*

The specification of monotonicity constraints is critical for ensuring that constrained models reflect economically meaningful relationships while maintaining predictive performance. We follow a systematic protocol to determine which features receive constraints and in what direction, based on four principles:

(i) Economic reasoning and domain knowledge.

(ii) Prior credit risk literature.

(iii) Train-only evidence to avoid test set leakage.

(iv) Encoding verification to ensure numeric order aligns with economic meaning.

Constraint signs follow a standard convention: +1 indicates an increasing constraint (higher feature value → higher default risk), while -1 indicates a decreasing constraint (higher feature value → lower default risk). Features with ambiguous or mixed evidence are left unconstrained, reflecting our conservative approach to constraint specification. This selective application ensures that constraints are imposed only where the economic relationship is clear and widely accepted in credit risk literature.

Table 2 presents a summary of constraint coverage across datasets. Coverage varies substantially, reflecting dataset characteristics and the selective application of constraints based on economic reasoning. Polish Bankruptcy has the highest coverage (41 features, predominantly decreasing), reflecting the extensive financial ratio constraints that are well-established in corporate bankruptcy literature. Taiwan Credit has 9 features (all increasing), consistent with payment history constraints where higher delay codes correspond to higher risk. Coverage ranges from 3 features (Lending Club) to 41 features (Polish Bankruptcy), demonstrating that constraint specification is dataset-specific rather than uniform.

TABLE 2 – MONOTONICITY CONSTRAINTS SUMMARY

| Dataset | Total Features | Constrained Features | Coverage (%) | Increasing (+1) | Decreasing (-1) |
|---|---|---|---|---|---|
| Taiwan Credit | 23 | 9 | 39.1 | 9 | 0 |
| Polish Bankruptcy | 64 | 41 | 64.1 | 5 | 36 |
| German Credit (Statlog) | 20 | 4 | 20.0 | 3 | 1 |
| Give Me Some Credit | 10 | 6 | 60.0 | 5 | 1 |
| Lending Club (Zenodo) | 8 | 3 | 37.5 | 1 | 2 |

*Notes*: Constraints were specified based on economic reasoning, domain knowledge, and prior credit risk literature. Constraint direction was verified using train-only evidence (no test set leakage) and encoding verification to ensure numeric order aligns with economic meaning. Increasing constraints (+1) indicate that higher feature values correspond to higher default risk (e.g., credit amount, debt-to-income ratio). Decreasing constraints (-1) indicate that higher feature values correspond to lower default risk (e.g., FICO score, profitability ratios). Features with ambiguous or mixed evidence were left unconstrained.

The complete list of all constrained features, including descriptions and economic reasoning for each constraint, is provided in Appendix A. Additional Tables (Table A.1). This table provides full transparency on constraint specification, enabling readers to evaluate the economic justification for each constraint. A visual summary of constraint coverage across datasets, showing the distribution of increasing versus decreasing constraints, is provided in Appendix B. Additional Figures – Figure B.1.

Constraint specification examples illustrate the economic reasoning underlying our approach. For credit amount and debt-to-income ratio, higher values indicate greater financial exposure or burden, leading to increasing constraints (+1). For FICO scores and income measures, higher values indicate better creditworthiness or repayment capacity, leading to decreasing constraints (-1). Payment delay indicators receive increasing constraints, as more delays correspond to worse payment history and higher default risk. Financial ratios in corporate datasets follow established bankruptcy prediction literature: profitability and efficiency ratios receive decreasing constraints, while leverage ratios receive increasing constraints.

### *E. Model Specifications*

We evaluate monotonicity constraints using three gradient boosting libraries: XGBoost (Chen and Guestrin 2016), LightGBM (Ke et al. 2017), and CatBoost (Prokhorenkova et al. 2018). All three support monotonicity constraints for numeric features, enabling direct comparison across implementations.

Each library enforces constraints during tree construction by restricting split decisions to maintain the specified monotonic relationships. The constraint specification format differs across libraries, but all accept increasing constraints (higher feature values correspond to higher default risk), decreasing constraints (higher feature values correspond to lower default risk), or no constraint. Constraints apply only to numeric features; categorical features are handled by each library's native encoding and are not subject to monotonicity constraints.

We maintain comparable base configurations across libraries. Hyperparameter search spaces are specified in Section III.F and applied identically to constrained and unconstrained models.

For datasets with categorical features (notably German Credit), LightGBM and CatBoost handle these natively. XGBoost requires explicit categorical type specification. Monotonicity constraints are applied only to numeric features, as categorical features lack a natural ordering. This reflects standard practice in credit risk modeling, where constraints are most commonly applied to continuous or ordinal numeric variables such as credit amounts, debt ratios, and payment history indicators.

### *F. Hyperparameter Tuning*

We employ grid search with identical search space sizes across all three libraries. For each dataset, we define hyperparameter grids with equal cardinality: $5^4 = 625$ combinations for most datasets (Taiwan Credit, Polish Bankruptcy, German Credit, Give Me Some Credit) and $4^4 = 256$ combinations for Lending Club due to its large sample size (Table 3).

Hyperparameter selection proceeds via 3-fold cross-validation on the training set, with models scored by negative LogLoss. The same grid search space is used for unconstrained and constrained models. Models are trained to the full number of estimators specified in the grid (no early stopping).

TABLE 3 – HYPERPARAMETER GRID SEARCH CONFIGURATION

| Dataset | XGBoost | LightGBM | CatBoost | Notes |
|---|---|---|---|---|
| Taiwan Credit | 625 | 625 | 625 | 5 values per hyperparameter |
| Polish Bankruptcy | 625 | 625 | 625 | 5 values per hyperparameter |
| German Credit (Statlog) | 625 | 625 | 625 | 5 values per hyperparameter |
| Give Me Some Credit | 625 | 625 | 625 | 5 values per hyperparameter |
| Lending Club (Zenodo) | 256 | 256 | 256 | 4 values per hyperparameter; reduced due to large dataset size |

*Notes*: Grid search performed using 3-fold cross-validation on the training set, scoring by negative LogLoss. Grid sizes are identical across libraries within each dataset. Lending Club uses a reduced grid (4 values per hyperparameter) due to computational constraints from the large dataset size (~1M samples).

For each library, we tune four key hyperparameters that control model complexity and regularization. XGBoost hyperparameters include learning rate, number of estimators, maximum tree depth, and L2 regularization. LightGBM hyperparameters include learning rate, number of estimators, number of leaves, and minimum child samples. CatBoost hyperparameters include learning rate, number of estimators, tree depth, and L2 leaf regularization. The complete grid search specifications are summarized in Table 3, and the best hyperparameters selected for each model are provided in Appendix A. Additional Tables.

### *G. Evaluation Metrics and Price of Monotonicity*

We evaluate model performance using four metrics that capture discrimination and calibration. The area under the receiver operating characteristic curve (AUC) measures discrimination – the model's ability to distinguish defaulters from non-defaulters across all classification thresholds (Hanley and McNeil 1982). The area under the precision-recall curve (PR-AUC) provides a complementary discrimination measure that is more informative under class imbalance, as it focuses on the positive class rather than the overall ranking (Davis and Goadrich 2006). Both AUC and PR-AUC are higher-is-better metrics, with values ranging from 0 to 1.

For calibration assessment, we use two proper scoring rules. The logarithmic loss (LogLoss) quantifies the penalty for predicted probabilities that deviate from the true binary outcomes, with lower values indicating better calibration (Gneiting and Raftery 2007). The Brier score measures the mean squared difference between predicted probabilities and actual outcomes, providing a direct assessment of probabilistic calibration (Brier 1950). Both LogLoss and Brier score are lower-is-better metrics.

To quantify the performance cost of monotonicity constraints, we define the Price of Monotonicity (PoM) as the relative change in each metric when moving from unconstrained to monotone-constrained models. This relative measure enables comparison across datasets with different baseline performance levels and facilitates interpretation of constraint effects in percentage terms.

Equations below provide the PoM formulas for both metric types:

For Higher-is-Better Metrics (AUC, PR-AUC):

$$PoM_{AUC} = \frac{AUC_{uncon} - AUC_{mono}}{AUC_{uncon}} \times 100\% \tag{1}$$

$$PoM_{PR-AUC} = \frac{PR-AUC_{uncon} - PR-AUC_{mono}}{PR-AUC_{uncon}} \times 100\% \tag{2}$$

For Lower-is-Better Metrics (LogLoss, Brier):

$$PoM_{LogLoss} = \frac{LogLoss_{mono} - LogLoss_{uncon}}{LogLoss_{uncon}} \times 100\% \tag{3}$$

$$PoM_{Brier} = \frac{Brier_{mono} - Brier_{uncon}}{Brier_{uncon}} \times 100\% \tag{4}$$

where $uncon$ denotes unconstrained models and $mono$ denotes monotone-constrained models.

The PoM interpretation is consistent across metrics: positive values indicate that monotonicity constraints impose a performance cost (lower discrimination for AUC/PR-AUC, worse calibration for LogLoss/Brier), negative values indicate that constraints improve performance, and values near zero indicate negligible differences. Uncertainty in PoM estimates is quantified using paired bootstrap resampling, as described in the next section.

*H. Statistical Inference: Paired Bootstrap*

We quantify uncertainty in Price of Monotonicity estimates using a paired bootstrap procedure. For each bootstrap replicate $b = 1, \ldots, B$ (where $B = 1{,}000$), we sample test set indices with replacement, ensuring both classes are present in the bootstrap sample. The same bootstrap sample is used to compute metrics for both unconstrained and constrained models, preserving the paired structure and accounting for correlation between model predictions. For each replicate $b$, we compute the Price of Monotonicity on the bootstrap sample:

$$PoM_{AUC}^{(b)} = \frac{AUC_{uncon}^{(b)} - AUC_{mono}^{(b)}}{AUC_{uncon}^{(b)}} \times 100\%$$

where $AUC_{uncon}^{(b)}$ and $AUC_{mono}^{(b)}$ denote the AUC computed on the same bootstrap sample for unconstrained and constrained models, respectively. The same procedure applies to PR-AUC, LogLoss, and Brier score, with PoM formulas adjusted for lower-is-better metrics as defined in Equations (1) – (4).

We report two summary statistics from the bootstrap distribution. The mean PoM is:

$$\overline{PoM}_{AUC} = \frac{1}{B} \sum_{b=1}^{B} PoM_{AUC}^{(b)}$$

We also report the 95% percentile confidence interval, defined as:

$$CI_{95\%} = [Q_{2.5\%}, Q_{97.5\%}]$$

where $Q_{\alpha}$ denotes the $\alpha$-th percentile of the bootstrap distribution of $PoM_{AUC}^{(b)}$ values. Specifically, $Q_{2.5\%}$ is the value below which 2.5% of the bootstrap PoM values fall, and $Q_{97.5\%}$ is the value below which 97.5% of the bootstrap PoM values fall. A confidence interval that excludes zero indicates a statistically detectable effect of monotonicity constraints on model performance.

## IV. Results

Monotonicity constraints impose modest performance costs on average – mean AUC PoM around 0.6–0.9% and mean Brier PoM around 2.6–3.2% – with larger effects in Polish Bankruptcy (2.4% AUC, 12.9% Brier) and almost no cost in the largest datasets (0.1–0.2% AUC PoM). We first look at these aggregate patterns and then turn to per-dataset results and diagnostics.

### *A. Overall Price of Monotonicity*

We examine the aggregate performance effects of monotonicity constraints across all datasets and libraries. Figure 1 displays the distributions of the Price of Monotonicity (PoM) for four key metrics: AUC, PR-AUC, LogLoss, and Brier score. The violin plots reveal substantial heterogeneity across dataset-library combinations, with most distributions centered near zero but exhibiting meaningful variation.

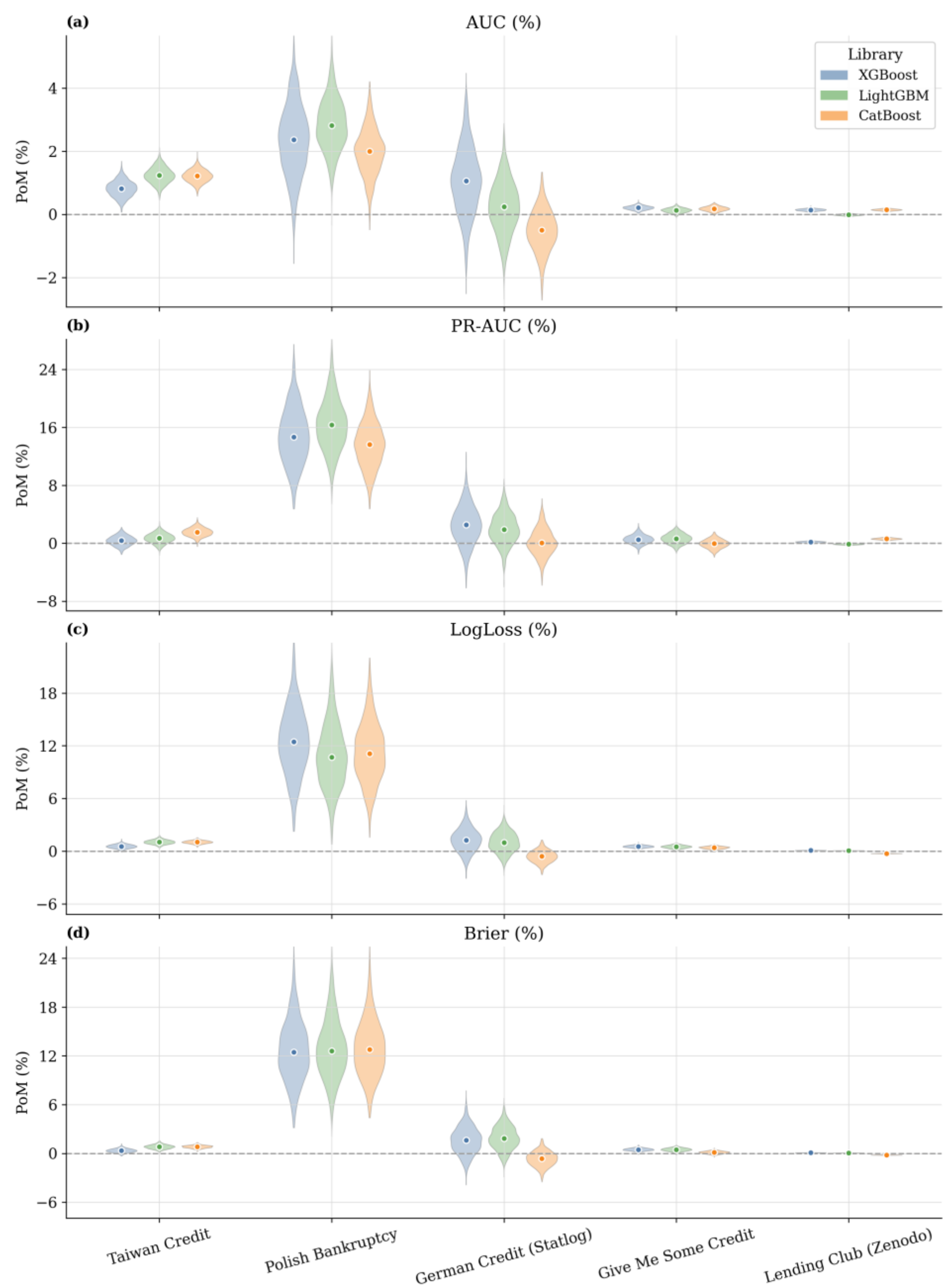

FIGURE 1. PRICE OF MONOTONICITY ACROSS DATASETS AND LIBRARIES: VIOLIN DISTRIBUTIONS (AUC, PR-AUC, LOGLOSS, BRIER)

*Notes*: Panels (a)–(d) show the distributions of the Price of Monotonicity (PoM, %) by dataset and library (colors). The horizontal dashed line marks zero PoM. Positive values indicate a performance cost from imposing monotonicity; negative values indicate a benefit. (a) AUC (%): higher is better; positive PoM means lower AUC under monotonicity. (b) PR-AUC (%): higher is better; positive PoM means lower PR-AUC under monotonicity. (c) LogLoss (%): lower is better; positive PoM means higher LogLoss under monotonicity. (d) Brier (%): lower is better; positive PoM means higher Brier under monotonicity. Distributions are computed across resamples; markers indicate central tendency for each violin.

Monotonicity constraints impose modest costs on discrimination metrics while showing more pronounced effects on calibration. The library-level average AUC PoM ranges from 0.6% (CatBoost) to 0.9% (XGBoost and LightGBM), with dataset-level averages ranging from 0.1% (Lending Club) to 2.4% (Polish Bankruptcy) (see Table 5 and Table 6). This suggests that the discrimination cost of monotonicity is generally small – often less than one percentage point – and in some cases effectively zero. The Brier score PoM, which captures calibration effects, shows larger variation: the library-level means range from 2.6% (CatBoost) to 3.2% (LightGBM), with a notable outlier in the Polish Bankruptcy dataset (12.9% mean Brier PoM across libraries).

Statistical significance, as measured by 95% paired-bootstrap confidence intervals that exclude zero, varies substantially by metric and library (Table 4). For AUC, 67% of dataset-library pairs show statistically significant PoM, with XGBoost and CatBoost each achieving significance in 80% of their dataset pairs, compared to 40% for LightGBM. PR-AUC shows lower significance rates overall (33%), with particularly low rates for XGBoost (20%) and LightGBM (20%), while CatBoost maintains 60% significance. Brier score significance is high across libraries (67% overall), with LightGBM showing the highest rate (80%).

TABLE 4 – SIGNIFICANCE OF PRICE OF MONOTONICITY BY METRIC AND LIBRARY (95% CI EXCLUDES ZERO)

| Metric | Significant (Total) | XGBoost | LightGBM | CatBoost |
|---|---|---|---|---|
| AUC | 10/15 (67%) | 4/5 (80%) | 2/5 (40%) | 4/5 (80%) |
| PR-AUC | 5/15 (33%) | 1/5 (20%) | 1/5 (20%) | 3/5 (60%) |
| Brier score | 10/15 (67%) | 3/5 (60%) | 4/5 (80%) | 3/5 (60%) |

*Notes*: “Significant” means the 95% paired-bootstrap CI for PoM excludes zero. Cells report count/total and percent across dataset × library pairs. Results reflect properly signed monotonicity constraints only; wrong-sign diagnostic results are summarized separately.

The library-level summary (Table 5) reveals that CatBoost exhibits the smallest average AUC PoM (0.6%) and Brier PoM (2.6%), though with substantial cross-dataset variation (SD of 1.0% for AUC, 5.8% for Brier). XGBoost and LightGBM show nearly identical average AUC PoM (0.9% each), but LightGBM displays slightly higher variance (SD of 1.2% versus 0.9% for XGBoost). The minimum AUC PoM values are close to zero across all libraries, with CatBoost showing a slight negative value (-0.5%), indicating a small benefit from constraints in at least one dataset. Maximum values are modest: 2.3% for XGBoost, 2.9% for LightGBM, and 2.0% for CatBoost.

TABLE 5 – LIBRARY-LEVEL SUMMARY OF PRICE OF MONOTONICITY (AUC, BRIER)

| Library | AUC PoM Mean (%) | AUC PoM SD (%) | AUC PoM Min (%) | AUC PoM Max (%) | Brier PoM Mean (%) | Brier PoM SD (%) | Brier PoM Min (%) | Brier PoM Max (%) |
|---|---|---|---|---|---|---|---|---|
| XGBoost | 0.9 | 0.9 | 0.1 | 2.3 | 3.0 | 5.4 | 0.1 | 12.7 |
| LightGBM | 0.9 | 1.2 | -0.0 | 2.9 | 3.2 | 5.5 | 0.1 | 12.9 |
| CatBoost | 0.6 | 1.0 | -0.5 | 2.0 | 2.6 | 5.8 | -0.7 | 13.0 |

*Notes*: "Mean" is the average across five datasets of each library's per-dataset PoM mean; "SD" is the cross-dataset standard deviation; "Min/Max" are the cross-dataset extrema.

Dataset-level patterns (Table 6) demonstrate that the cost of monotonicity varies meaningfully across datasets. Polish Bankruptcy shows the highest average AUC PoM (2.4%) and dramatically higher Brier PoM (12.9%), suggesting that this dataset's feature structure makes monotonicity constraints particularly costly. Conversely, Lending Club exhibits minimal costs (0.1% AUC PoM, 0.0% Brier PoM mean), indicating that monotonicity is essentially free for this dataset. Taiwan Credit shows moderate costs (1.1% AUC PoM), while German Credit and Give Me Some Credit show very small effects (0.3% and 0.2% AUC PoM, respectively).

TABLE 6 – DATASET-LEVEL SUMMARY OF PRICE OF MONOTONICITY (AUC, BRIER)

| Dataset | AUC PoM Mean (%) | AUC PoM SD (%) | AUC PoM Min (%) | AUC PoM Max (%) | Brier PoM Mean (%) | Brier PoM SD (%) | Brier PoM Min (%) | Brier PoM Max (%) |
|---|---|---|---|---|---|---|---|---|
| Taiwan Credit | 1.1 | 0.2 | 0.8 | 1.2 | 0.7 | 0.3 | 0.3 | 0.8 |
| Polish Bankruptcy | 2.4 | 0.4 | 2.0 | 2.9 | 12.9 | 0.2 | 12.7 | 13.0 |
| German Credit (Statlog) | 0.3 | 0.8 | -0.5 | 1.1 | 0.9 | 1.4 | -0.7 | 1.9 |
| Give Me Some Credit | 0.2 | 0.0 | 0.1 | 0.2 | 0.4 | 0.2 | 0.1 | 0.5 |
| Lending Club (Zenodo) | 0.1 | 0.1 | 0.0 | 0.1 | 0.0 | 0.2 | -0.2 | 0.1 |

*Notes*: "Mean" is the average across libraries of each dataset's per-library PoM mean; "SD" is the cross-library standard deviation; "Min/Max" are the cross-library extrema.

The heterogeneity in PoM across datasets reflects differences in the alignment between the data-generating process and the imposed constraints. When the underlying relationships are naturally monotone, constraints impose little cost and may even improve performance through variance reduction. When monotonicity conflicts with the true relationships – or when the constrained features are particularly informative – the cost rises. The Polish Bankruptcy dataset's high Brier PoM (12.9%) suggests that monotonicity constraints may be forcing the model away from optimal calibration in ways that do not fully trade off against discrimination gains.

These aggregate results establish that monotonicity constraints are not uniformly costly: in many cases, the discrimination cost is negligible, while calibration effects vary more substantially. The next subsection examines dataset-specific patterns to understand when and why these costs emerge.

### *B. Dataset-Specific Results*

We now examine dataset-specific patterns to understand when and why monotonicity constraints impose costs or provide benefits. Table 7 reports the Price of Monotonicity for each dataset-library combination across all metrics. The results reveal substantial heterogeneity that reflects differences in feature structure, data characteristics, and the alignment between imposed constraints and underlying relationships.

TABLE 7 — PRICE OF MONOTONICITY (POM) BY DATASET AND LIBRARY

| Dataset | Model library | PoM – (AUC, %) [95% CI] | PoM – (PR-AUC, %) [95% CI] | PoM – (Brier score, %) [95% CI] | Test set size, $n$ |
|---|---|---|---|---|---|
| Taiwan Credit | XGBoost | **0.8 [0.3, 1.3]** | 0.3 [-0.9, 1.6] | 0.3 [-0.1, 0.8] | 9,000 |
| | LightGBM | **1.2 [0.8, 1.7]** | 0.7 [-0.7, 2.0] | **0.8 [0.4, 1.3]** | |
| | CatBoost | **1.2 [0.8, 1.6]** | **1.5 [0.4, 2.7]** | **0.8 [0.5, 1.2]** | |
| Polish Bankruptcy | XGBoost | **2.3 [0.1, 4.4]** | **14.8 [7.9, 22.4]** | **12.7 [5.8, 20.2]** | 3,151 |
| | LightGBM | **2.9 [1.3, 4.6]** | **16.4 [9.8, 24.1]** | **12.9 [6.5, 21.1]** | |
| | CatBoost | **2.0 [0.6, 3.5]** | **13.6 [7.8, 19.7]** | **13.0 [6.8, 20.4]** | |
| German Credit (Statlog) | XGBoost | 1.1 [-1.0, 3.2] | 2.5 [-2.6, 7.6] | 1.6 [-1.4, 4.7] | 300 |
| | LightGBM | 0.2 [-1.3, 1.8] | 2.0 [-2.6, 6.2] | 1.9 [-0.9, 4.7] | |
| | CatBoost | -0.5 [-1.9, 0.8] | 0.1 [-3.3, 3.9] | -0.7 [-2.4, 1.1] | |
| Give Me Some Credit | XGBoost | **0.2 [0.1, 0.3]** | 0.5 [-0.6, 1.7] | **0.5 [0.2, 0.8]** | 45,000 |
| | LightGBM | 0.1 [-0.0, 0.3] | 0.6 [-0.7, 2.0] | **0.5 [0.1, 0.8]** | |
| | CatBoost | **0.2 [0.0, 0.3]** | -0.1 [-1.2, 1.1] | 0.1 [-0.2, 0.4] | |
| Lending Club (Zenodo) | XGBoost | **0.1 [0.1, 0.2]** | 0.2 [-0.0, 0.4] | **0.1 [0.0, 0.1]** | 225,277 |
| | LightGBM | -0.0 [-0.1, 0.0] | -0.1 [-0.3, 0.0] | **0.1 [0.0, 0.1]** | |
| | CatBoost | **0.1 [0.1, 0.2]** | **0.6 [0.4, 0.8]** | -0.2 [-0.2, -0.2] † | |

*Notes*: Bold indicates the 95% CI excludes zero and PoM > 0 (cost). † Indicates the 95% CI excludes zero and PoM < 0 (benefit). With our PoM definitions, negative values imply a benefit for both AUC and Brier.

*Source*: Author calculations.

**Taiwan Credit** shows moderate costs, with average AUC PoM of 1.1% and Brier PoM of 0.7% (Table 6). XGBoost exhibits the smallest AUC PoM (0.8%) with a non-significant Brier PoM (0.3%, CI includes zero), while LightGBM and CatBoost show larger but still modest costs (1.2% AUC PoM and 0.8% Brier PoM, both significant) (Table 7). The relatively small calibration cost compared to Polish Bankruptcy suggests that constraints align better with the true relationships in this dataset, which has 23 features capturing credit card repayment history, bill amounts, and payment amounts.

**Give Me Some Credit** and **Lending Club** exhibit minimal costs, with average AUC PoM of 0.2% and 0.1%, respectively (Table 6). Give Me Some Credit shows 0.4% Brier PoM, while Lending Club demonstrates essentially zero Brier PoM (0.0%). In Lending Club, CatBoost even shows a small calibration benefit (-0.2% Brier PoM, significant). Both datasets have fewer features (10 and 8, respectively) focused on core credit risk factors. Lending Club uses a temporal split (train 2007–2015, test 2017–2018), while Give Me Some Credit uses a stratified split.

**German Credit** presents an interesting pattern: while the average AUC PoM is small (0.3%), results vary dramatically across libraries. XGBoost shows 1.1% AUC PoM, but the 95% CI includes zero, indicating statistical uncertainty. LightGBM exhibits 0.2% AUC PoM (CI includes zero). Most notably, CatBoost shows negative PoM for both AUC (-0.5%) and Brier (-0.7%), suggesting benefits from constraints (Table 7). This dataset has the smallest test set (300 observations), which contributes to wide confidence intervals, making statistical significance harder to detect. The dataset contains 20 features with a mix of numeric and categorical variables.

The dataset-specific results reveal several key patterns. The cost of monotonicity varies substantially across datasets, from essentially zero (Lending Club) to substantial (Polish Bankruptcy), reflecting differences in feature complexity, data characteristics, and the alignment between constraints and underlying relationships. Library differences matter: CatBoost shows the smallest costs in some datasets (Polish Bankruptcy, German Credit) and occasionally provides benefits (German Credit, Lending Club Brier), while XGBoost and LightGBM show more variable patterns across datasets. Calibration costs (Brier PoM) can be much larger than discrimination costs (AUC PoM), particularly when constraints conflict with optimal probability calibration, as seen in Polish Bankruptcy. Small datasets (German Credit) show wider confidence intervals, making it harder to detect statistically significant effects even when point estimates suggest benefits or costs.

### *C. Diagnostic Analyses*

To validate that monotonicity constraints meaningfully affect model performance and that the chosen constraint signs are economically appropriate, we conduct a diagnostic analysis using intentionally mis-specified constraints. We train XGBoost models with wrong-sign monotonicity constraints – constraints that enforce the opposite direction of the economically justified signs – and compare their performance to both unconstrained and correctly constrained models.

Figure B.2 (Appendix B. Additional Figures) compares PoM distributions for correctly specified versus mis-specified constraints (XGBoost only), showing that correctly specified constraints consistently produce lower PoM across all metrics and datasets.

Figure 2 displays the Price of Monotonicity for wrong-sign constrained models across all datasets. The results demonstrate that mis-specified constraints impose substantially larger costs than correctly specified constraints. On average across datasets, wrong-sign constraints produce an AUC PoM of 8.6% and a Brier PoM of 11.0%, compared to 0.9% and 3.0% respectively for correctly specified constraints (XGBoost). This represents approximately a 10-fold increase in discrimination cost and a 4-fold increase in calibration cost when constraints are mis-specified.

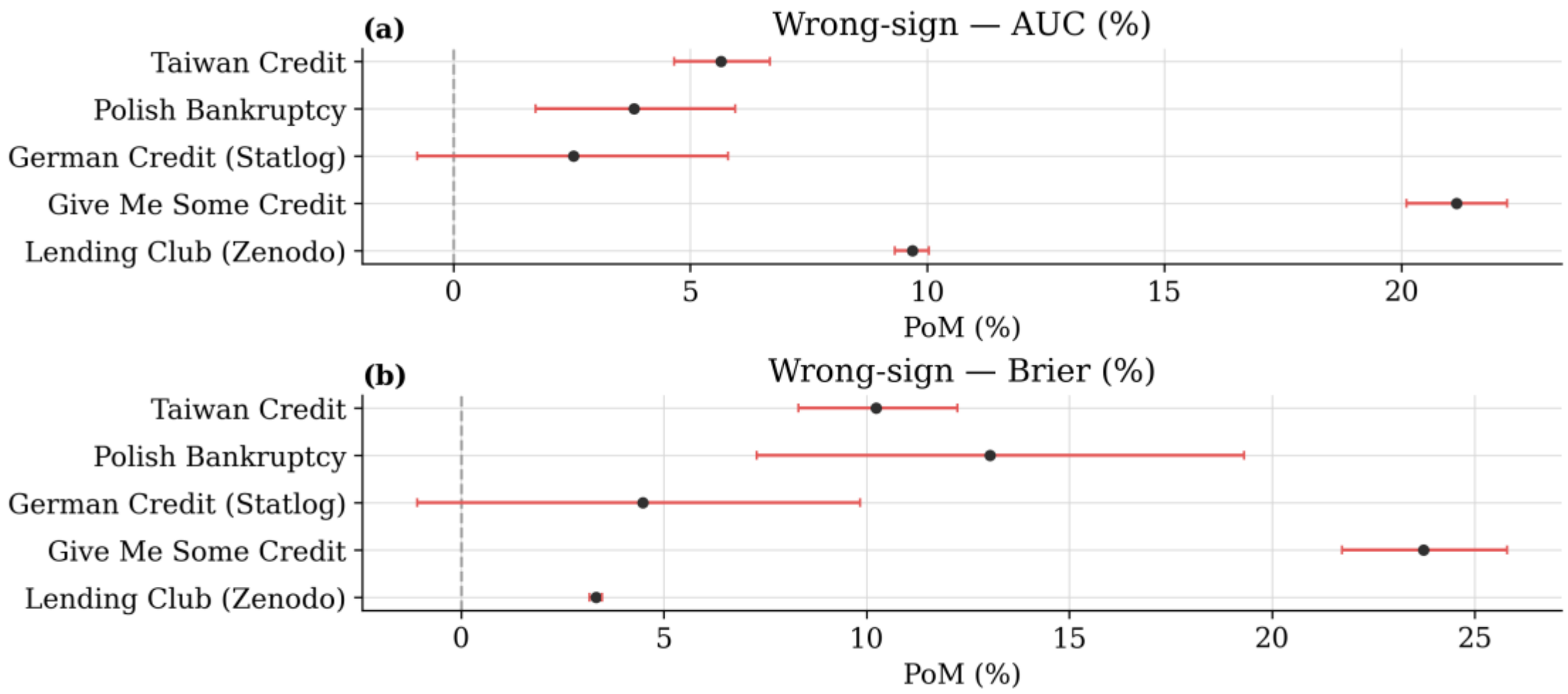

FIGURE 2. WRONG-SIGN MONOTONICITY CONSTRAINTS: PRICE OF MONOTONICITY FOR AUC AND BRIER BY DATASET

*Notes*: Panels (a)–(b) report the PoM (%) for intentionally wrong-sign constrained models relative to unconstrained, by dataset. Dots show mean PoM; horizontal lines show 95% CIs from paired bootstrap; the vertical dashed line marks zero PoM. (a) AUC (%): higher is better; positive PoM indicates an AUC loss under wrong-sign constraints (cost), negative indicates a benefit. (b) Brier (%): lower is better; positive PoM indicates higher Brier (worse calibration) under wrong-sign constraints, negative indicates improved calibration.

The magnitude of wrong-sign costs varies substantially across datasets, reflecting differences in the informativeness of constrained features and the strength of the true monotonic relationships. Give Me Some Credit shows the largest effects: wrong-sign constraints produce an AUC PoM of 21.2% (101 times larger than the correct-sign cost of 0.2%) and a Brier PoM of 23.7% (49 times larger than the 0.5% correct-sign cost). Lending Club also exhibits large wrong-sign costs (9.7%

AUC PoM, 68 times the correct-sign cost), while Taiwan Credit shows a 7-fold increase in AUC PoM (5.6% versus 0.8%) and a 30-fold increase in Brier PoM (10.2% versus 0.3%). In contrast, Polish Bankruptcy – which already shows high costs under correct constraints – exhibits more modest wrong-sign effects (1.6-fold increase in AUC PoM), suggesting that this dataset's feature structure makes it inherently costly to constrain regardless of sign direction.

These diagnostic results serve two important purposes. First, they confirm that monotonicity constraints have meaningful effects on model performance: the large costs observed under wrong-sign constraints demonstrate that the constraints are binding and that the models cannot simply ignore them. Second, they validate the economic reasoning underlying the constraint signs: the dramatic difference between correct-sign and wrong-sign costs provides evidence that the chosen constraint directions align with the true relationships in the data. The fact that wrong-sign constraints consistently produce larger costs than correct-sign constraints – often by an order of magnitude or more – suggests that the constraint specification protocol successfully identified economically meaningful monotonic relationships.

Statistical significance is high for wrong-sign constraints: 80% of dataset-metric pairs show statistically significant PoM (4 of 5 datasets for both AUC and Brier), with all significant results showing positive PoM (performance costs). The one exception is German Credit, where wrong-sign constraints show non-significant effects across all metrics despite positive point estimates (AUC PoM: 2.5%, Brier PoM: 4.5%). This likely reflects the dataset's small test set size (~300 observations), which leads to wider bootstrap confidence intervals and reduced statistical power. This contrasts with the more mixed significance patterns observed for correctly specified constraints, where some datasets show negligible or statistically insignificant effects. The consistent and large wrong-sign costs provide a useful benchmark: when constraints impose costs similar to or larger than wrong-sign costs, it may indicate that the constraint specification needs reconsideration.

Figure 3 illustrates the effect of constraints on individual feature relationships using Individual Conditional Expectation (ICE) and Partial Dependence Plot (PDP) curves for selected features. The plots compare unconstrained and constrained XGBoost models, showing how constraints enforce smooth monotonic relationships while unconstrained models can exhibit local non-monotonicities. For Taiwan Credit's repayment status feature (X6), constraints enforce a non-decreasing relationship that suppresses spurious dips visible in the unconstrained model. For Lending Club's FICO score, constraints enforce a non-increasing relationship (higher FICO scores correspond to lower default risk) and reduce local wiggles. These visualizations complement the aggregate PoM results by demonstrating how constraints affect model behavior at the feature level.

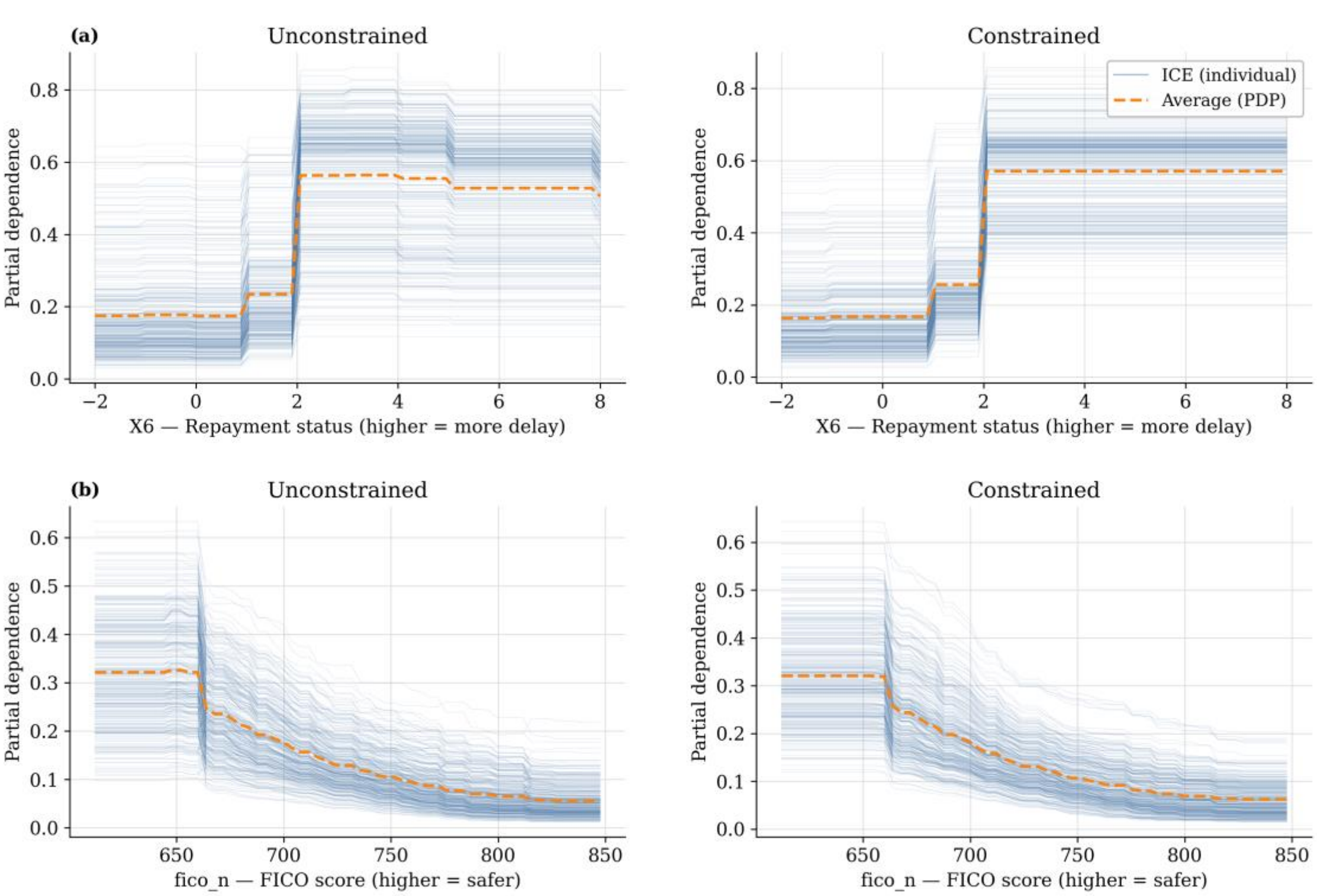

Figure 3. Selected ICE/PDP under Unconstrained vs Monotone-Constrained XGBoost (Taiwan: « Repayment status»; Lending Club: «FICO»)

*Notes*: Columns compare Unconstrained (left) and Monotone-Constrained (right) XGBoost. Thin blue lines are ICE curves for randomly sampled individuals; the orange dashed line is the PDP (average). (a) Taiwan Credit — feature X6, repayment status (higher = more delay): constraints enforce a non-decreasing relationship and suppress spurious dips visible in the unconstrained model. (b) Lending Club – fico_n, FICO score (higher = safer): constraints enforce a non-increasing relationship and reduce local wiggles. Probabilities are raw model outputs; X6 is integer-coded, hence step-like PDP segments.

Together, these diagnostic analyses indicate that monotonicity constraints do affect model behavior and are consistent with the chosen constraint signs being economically appropriate. The large costs observed under wrong-sign constraints support the constraint specification protocol, while the ICE/PDP visualizations illustrate that constraints enforce the intended monotonic relationships at the feature level.

## V. Discussion

The Price of Monotonicity is small on average but varies across datasets, libraries, and constraint specifications, so simple "always expensive" or "always free" rules don't hold. We discuss when constraints are effectively free versus costly, how implementations differ, how calibration and discrimination trade off, practical recommendations, and limitations.

### *A. When Monotonicity is Free vs. Costly*

The performance cost of monotonicity varies substantially across datasets: in some, constraints are essentially "free", in others they entail measurable discrimination losses. Understanding when constraints are free versus costly has important implications for practitioners deciding whether to incorporate domain knowledge into their models.

A key pattern in our results is the relationship between dataset size and constraint cost. On the two largest datasets – Give Me Some Credit (n ≈ 150,000) and Lending Club (n ≈ 1,000,000) – the Price of Monotonicity for AUC is consistently small, typically below 0.2% and often not distinguishable from zero. For LightGBM on Lending Club, the constrained model achieves a slightly lower (though not significantly different) AUC PoM of -0.02%, suggesting that in this context, monotonicity constraints may function as a form of regularization that reduces overfitting without sacrificing discrimination. This pattern aligns with theoretical expectations: when data are abundant, constraints that align with the true data-generating process can reduce model variance by preventing the algorithm from fitting spurious patterns, potentially offsetting any bias introduced by the constraint (Hastie et al. 2009; Bzdok 2017).

Conversely, the highest costs appear on Polish Bankruptcy, where AUC PoM ranges from approximately 2.0% to 2.9% across libraries, with all estimates statistically significant. This dataset is notable for having the largest number of constrained features (41 out of 64 total features) relative to its sample size ($n \approx 10{,}500$). The substantial constraint coverage – approximately 64% of features—combined with a moderate sample size may create a scenario where the constraints collectively restrict model flexibility more than the data can support. This pattern suggests that the ratio of constrained features to sample size, rather than absolute sample size alone, may be an important determinant of constraint cost.

The German Credit dataset presents an interesting intermediate case. With only 1,000 observations and 4 constrained features (20% of features), the costs are generally small and often not statistically significant. Notably, CatBoost achieves a negative AUC PoM of -0.5% (though the confidence interval includes zero), again suggesting potential regularization benefits. The small number of constraints relative to the dataset's dimensionality may allow the model sufficient flexibility to capture important patterns while still benefiting from the variance-reducing effects of constraints on the constrained features.

These patterns hint at a potential mechanism through which monotonicity constraints affect performance. When constraints align well with the underlying data-generating process and the model retains sufficient flexibility on unconstrained features, constraints may act as a form of structural regularization. This interpretation is consistent with work on shape-constrained Gaussian processes, which shows that incorporating monotonicity or related inequality constraints can substantially reduce posterior uncertainty while leaving predictive accuracy largely unchanged (Agrell 2019), and that misspecified monotonicity assumptions may introduce bias and degrade accuracy in some settings (Riihimäki and Vehtari 2010). Similar regularization benefits from monotonicity have also been documented in deep lattice networks, where monotone constraints improve generalization while enhancing interpretability (You et al. 2017).

The wrong-sign diagnostic provides additional insight into the cost of misspecification. When constraints are applied with incorrect signs (e.g., constraining income to increase default risk), the costs are substantial – ranging from approximately 2.5% to 21.2% AUC PoM across datasets. This suggests that the cost of constraints depends critically on whether they align with the true relationships in the data. When constraints are well-specified based on domain knowledge, costs are generally small; when they are mis-specified, costs can be substantial.

Several factors may help explain when monotonicity is more likely to be "free". First, large sample sizes appear to mitigate costs, possibly because abundant data allow the model to learn complex patterns on unconstrained features while still benefiting from constraint-induced variance reduction. Second, constraining a smaller proportion of features may preserve model flexibility, allowing the algorithm to capture important interactions and non-monotonic effects on unconstrained features. Third, when constraints align closely with the true data-generating process – as suggested by domain knowledge and validated through diagnostics like ICE plots – the bias introduced may be minimal relative to the variance reduction achieved.

Conversely, constraints may be more costly when:

(i) Many features are constrained relative to sample size, reducing overall model flexibility.

(ii) Constraints conflict with important patterns in the data (though our domain-knowledge-based approach aims to minimize this).

(iii) The dataset is of moderate size, where the bias-variance trade-off is less favorable to constraint-induced regularization.

These findings have practical implications. For practitioners working with large datasets and clear domain knowledge about a subset of features, monotonicity constraints appear to offer interpretability benefits with minimal performance cost. However, when sample sizes are moderate and many features are constrained, practitioners should be prepared for potentially measurable discrimination losses and should carefully validate that constraints align with the data through diagnostic tools like partial dependence plots.

### *B. Library Differences and Implementation Effects*

Our results show modest but meaningful differences in the Price of Monotonicity across the three gradient boosting libraries. While the average AUC PoM is similar between XGBoost and LightGBM (0.9% each), CatBoost has a slightly lower average (0.6%), with cross-dataset standard deviations of 0.9%, 1.2%, and 1.0% respectively. Notably, CatBoost and LightGBM occasionally achieve negative PoM values – indicating a small benefit from constraints – in certain datasets (German Credit and Lending Club), while XGBoost consistently shows non-negative PoM across all datasets in our study.

These differences may reflect implementation-level variation in how monotonic constraints are enforced during tree construction. XGBoost uses a level-wise tree growth strategy with approximate split finding, LightGBM uses a leaf-wise (best-first) growth strategy with histogram-based optimization, and CatBoost implements ordered boosting with symmetric (oblivious) tree structures. These architectural differences could interact with monotonic constraint enforcement and thereby affect the trade-off between constraint satisfaction and predictive performance, although our results suggest that cross-dataset variation is larger than systematic cross-library differences.

The observation that CatBoost occasionally achieves negative PoM – indicating that constraints can sometimes improve rather than harm performance – is consistent with its ordered boosting mechanism, which reduces overfitting through permutation-based training. When constraints align with the underlying data-generating process, this regularization effect may complement the monotonicity constraint and yield modest gains in generalization. However, this pattern is not consistent across datasets, and the magnitudes are small (typically less than 1% in absolute terms), indicating that such benefits are context-dependent rather than systematic.

The cross-dataset variation in library-specific PoM patterns further suggests that implementation details interact with dataset characteristics: for example, in Polish Bankruptcy – where all libraries show elevated PoM (around 2.0% to 2.9% for AUC) – the high-dimensional financial ratio structure may amplify the impact of constraints, whereas in Lending Club, where PoM values are near zero across libraries, the constraints impose minimal cost regardless of implementation.

These findings align with broader evidence from other domains that library choice can meaningfully affect machine learning outcomes, even when algorithms are conceptually similar. Moussa and Sarro (2022) compare Scikit-Learn, Caret, and Weka implementations of the same deterministic learners for software effort estimation and find that the predictions produced by different libraries differ in 95% of the 105 cases they study, often by large margins. While our observed differences are more modest – reflecting the narrower scope of comparing monotonicity implementations within gradient boosting – they nonetheless underscore that implementation details matter when evaluating the cost of monotonicity.

Several limitations should be noted in interpreting these library differences. First, we use fixed, symmetric hyperparameter grids with equal cardinality across libraries, rather than library-specific search spaces that may be individually optimal; if the performance frontier differs by implementation, this could affect the relative PoM. Second, the constraint enforcement mechanisms are not directly observable in our analysis; we infer implementation differences from performance outcomes rather than from direct inspection of the underlying algorithms. Third, our results reflect specific versions of XGBoost, LightGBM, and CatBoost, and may not fully generalize to future implementations as these libraries evolve.

For practitioners, these findings suggest that library choice may matter when monotonicity constraints are a priority, though the differences are generally modest relative to the overall PoM. CatBoost's slightly lower average PoM and occasional negative values might make it attractive when constraints are essential, but the cross-dataset variation indicates that no single library consistently outperforms others. The more substantial differences emerge at the dataset level rather than the library level, reinforcing that the cost of monotonicity depends primarily on the alignment between constraints and the data structure rather than implementation details.

### *C. Calibration vs. Discrimination Trade-offs*

The relationship between calibration and discrimination is central to probability prediction. Discrimination, typically measured by the AUC, captures a model's ability to rank observations by risk. Calibration, often assessed with the Brier score, reflects the agreement between predicted probabilities and observed event rates. The Brier score can be decomposed into components associated with calibration (or reliability) and sharpness (or resolution) of the forecasts (Murphy 1973; DeGroot and Fienberg 1983), where the sharpness/refinement component captures how strongly the predicted probabilities are dispersed across the probability scale, conditional on calibration.

Monotonicity constraints may affect both calibration and discrimination through their regularizing role. By restricting the functional form, constraints can lower variance and improve the reliability of probability forecasts, but they may also limit the model's ability to capture complex interactions that contribute to high discrimination. Empirically, the net effect on these two dimensions is not straightforward.

Our results reveal heterogeneous patterns across datasets: discrimination losses are generally small, with the largest effects in Polish Bankruptcy. The calibration picture is more nuanced. In Polish Bankruptcy, constraints are associated with higher Brier scores (PoM 12–13%), suggesting a trade-off where discrimination losses coincide with calibration deterioration. Yet in other datasets, this pattern does not consistently hold. For German Credit, CatBoost shows a negative Brier PoM of -0.67% (95% CI: -2.45, 1.09), indicating potential calibration improvement, though the confidence interval includes zero. Similarly, on Lending Club, CatBoost exhibits a negative Brier PoM of -0.20% (95% CI: -0.24, -0.17), suggesting constraints may improve calibration in some contexts.

These mixed findings align with research suggesting that calibration and discrimination can trade off but need not always do so (Niculescu-Mizil and Caruana 2005). Recent work on multi-distribution learning has demonstrated that even at Bayes optimality, there exists an inherent calibration-refinement trade-off (Verma et al. 2024). The dataset-specific nature of these effects may reflect differences in the alignment between the unconstrained model's learned patterns and the imposed monotonicity constraints. When constraints align well with the underlying data-generating process, they may act as beneficial regularization, potentially improving both calibration and discrimination. When constraints conflict with important non-monotonic relationships, both aspects may suffer.

The absence of a consistent trade-off pattern suggests that the calibration-discrimination relationship under monotonicity constraints is context-dependent. This observation has practical implications: practitioners should evaluate both metrics when considering monotonicity constraints, rather than assuming that discrimination losses necessarily imply calibration gains, or vice versa. Future work could explore whether certain dataset characteristics – such as feature dimensionality, sample size, or the strength of monotonic relationships in the data – systematically predict when constraints improve or harm calibration.

### *D. Practical Recommendations*

Based on our empirical findings across five credit risk datasets, we offer several practical recommendations for practitioners considering monotonicity constraints in PD modeling. These suggestions reflect the heterogeneous nature of constraint effects observed in our study and aim to guide decision-making when interpretability and regulatory compliance are priorities.

**Feature selection and constraint specification.** Our results underscore the importance of careful feature selection when imposing monotonicity constraints. In our diagnostic wrong-sign experiment, mis-specified constraints generate large performance losses, with AUC PoM ranging from about 2.5% to over 20% across datasets. This highlights the risk of constraining features in directions that are not economically defensible. Practitioners should therefore constrain only those features for which there is a clear, well-supported monotonic relationship, grounded in domain knowledge and economic reasoning. Features with ambiguous or potentially non-monotonic effects – for example, age, which often exhibits non-linear or U-shaped patterns – are better left unconstrained unless monotonicity is supported by diagnostic tools such as ICE/PDP analyses or targeted sensitivity checks. The constraint selection protocol used in this study, which combines economic logic with literature evidence and explicitly excludes ambiguous features, can serve as a practical template for similar credit risk applications.

**Diagnostic validation and monitoring.** Given the context-dependent nature of constraint effects, practitioners should employ diagnostic tools to validate the appropriateness of monotonicity assumptions both before and after model deployment. In the ex ante stage, unconstrained partial dependence plots (PDPs) and individual conditional expectation (ICE) curves (interpreted with care in the presence of correlated predictors) can flag features where a strictly monotone relationship is implausible. After fitting the constrained model, comparing constrained PDPs/ICE curves to their unconstrained counterparts can reveal systematic distortions in feature effects. When constraints cut across important non-monotonic structure in the data, both discrimination and calibration may suffer—as in Polish Bankruptcy, where extensive constraints coincided with elevated PoM for AUC and Brier score. Embedding these diagnostics into model validation as a routine "monotonicity check" helps identify features for which constraints should be relaxed or removed, reducing the risk of performance losses when constraints misalign with the data-generating process.

**Multi-metric evaluation and library choice.** Our results suggest that practitioners should assess both discrimination and calibration when studying the impact of monotonicity constraints, rather than assuming a stable trade-off pattern. Calibration responses are notably heterogeneous: Brier PoM ranges from about –0.7% (German Credit, CatBoost) to 13.0% (Polish Bankruptcy, CatBoost), indicating that calibration effects are context-dependent and cannot be inferred from changes in discrimination alone. When monotonicity is required for regulatory or interpretability reasons, library choice appears to be a second-order consideration. CatBoost has a slightly lower average AUC PoM (0.6% vs. 0.9% for XGBoost and LightGBM) and occasionally yields negative PoM, which may make it an appealing default. However, variation across datasets is substantial and no library dominates uniformly, so practitioners should benchmark several implementations on their own data rather than assume universal superiority.

**Context-specific decision framework.** The substantial variation in PoM across datasets (AUC PoM from effectively 0% up to 2.9%) suggests that, in our benchmark, the cost of monotonicity is driven more by dataset characteristics and the alignment between constraints and data structure than by differences across implementations. Practitioners working with high-dimensional financial ratio datasets similar to Polish Bankruptcy should be prepared for potentially larger performance costs (AUC PoM around 2.0–2.9%, Brier PoM around 13%), whereas in large consumer credit datasets constraints can be almost costless (AUC PoM near zero, as in Lending Club). When preserving discrimination and calibration is critical, a pragmatic approach is to constrain only the most economically defensible features rather than applying constraints broadly, which may reduce the overall PoM while retaining interpretability benefits for key risk drivers.

### *E. Limitations and Future Directions*

Several limitations of this study should be acknowledged to contextualize our findings. First, our analysis is restricted to five public credit risk datasets, which may not fully capture the heterogeneity of production environments where PD models are deployed. While these datasets span different contexts (consumer credit, corporate bankruptcy, peer-to-peer lending), they represent curated, research-oriented collections that may differ from proprietary datasets in terms of feature engineering, data quality, and temporal dynamics. The generalizability of our PoM estimates to production settings remains an open question that could be addressed through replication studies using proprietary datasets or industry benchmarks.

Second, our evaluation is limited to tree-based gradient boosting algorithms (XGBoost, LightGBM, CatBoost). While these represent the dominant class of models used in credit risk modeling, other interpretable model families with monotonicity capabilities exist, including generalized additive models (GAMs) with monotone splines and monotone neural networks (Pya and Wood 2015; Meyer 2018; Sill 1997). The relative performance costs of monotonicity constraints may differ across model classes, and future work could compare PoM across these alternatives to provide a more comprehensive benchmark. Additionally, our constraint specification relies on domain knowledge and economic reasoning, which introduces an element of subjectivity. While we validated constraints using ICE/PDP diagnostics, automated methods for constraint discovery – such as testing monotonicity assumptions on training data or using statistical tests – could reduce this subjectivity and potentially identify constraints that improve both interpretability and performance.

Third, our study focuses exclusively on binary classification tasks in the credit risk domain. The cost of monotonicity constraints may differ for regression tasks (e.g., loss given default prediction) or multi-class problems, and extending the PoM framework to these settings would broaden its applicability. Similarly, our constraint selection protocol primarily targets numeric features, as categorical features require careful encoding to ensure meaningful monotonicity. While modern boosting libraries support categorical features natively, the interaction between categorical encoding strategies and monotonicity constraints remains underexplored and represents a promising direction for future research.

Fourth, our evaluation uses a single holdout split per dataset (or chronological split for temporal data), which provides a clean, regulator-friendly evaluation protocol but may not fully capture variance across different data partitions. While we employ paired bootstrap to quantify uncertainty, future work could assess PoM stability across multiple train-test splits or use cross-validation to provide more robust estimates of constraint effects.

Finally, our study examines constraints at the feature level, where each feature is either fully constrained or unconstrained. In practice, practitioners might benefit from more nuanced approaches, such as constraining only specific ranges of feature values or applying constraints with varying strength. Research on adaptive or soft monotonicity constraints could provide more flexible tools for balancing interpretability and performance.

Future research directions that could build upon these findings include:

(i) Extending the PoM benchmark to proprietary datasets and production environments to assess external validity.

(ii) Comparing monotonicity costs across model classes, particularly GAMs and monotone neural networks, which may offer different trade-offs.

(iii) Developing automated constraint selection methods that combine statistical testing with domain knowledge.

(iv) Exploring monotonicity in regression and multi-class settings.

(v) Investigating categorical feature encoding strategies that preserve meaningful monotonicity.

(vi) Developing adaptive constraint methods that allow partial or range-specific monotonicity.

Such extensions would strengthen the evidence base for when and how monotonicity constraints should be applied in credit risk modeling.

## VI. Conclusion

This paper benchmarks the Price of Monotonicity (PoM) for gradient boosting credit PD models across five public datasets and three libraries. In this setting, appropriately specified constraints rarely impose large discrimination costs: AUC PoM ranges from effectively zero to about 2.9% and is often near zero in large portfolios.

PoM provides a simple percentage-scale measure of the performance trade-off between unconstrained and monotone-constrained models. Because we estimate PoM using paired comparisons under identical experimental conditions, the differences we observe reflect constraint effects rather than design artifacts, giving practitioners and regulators a concrete sense of how much accuracy is typically sacrificed for monotonicity.

The most striking finding from our analysis is the conditional nature of constraint costs. On large datasets with abundant observations – Give Me Some Credit ($n \approx 150{,}000$) and Lending Club ($n \approx 1{,}000{,}000$) – monotonicity constraints impose minimal or negligible performance penalties, with AUC PoM typically below 0.2% and often statistically indistinguishable from zero. In some cases, constraints appear to function as a form of structural regularization, with LightGBM on Lending Club achieving a slightly negative PoM (-0.02%), suggesting that constraints can enhance rather than hinder performance when they align with the underlying data-generating process. This pattern aligns with theoretical expectations from bias-variance decomposition: when data are abundant, constraints that align with true relationships can reduce model variance by preventing overfitting to spurious patterns, potentially offsetting any bias introduced by the constraint.

Conversely, the highest costs emerge on Polish Bankruptcy, where AUC PoM ranges from 2.0% to 2.9% across all three libraries, with all estimates statistically significant. This dataset is notable for having the largest constraint coverage (41 out of 64 features, approximately 64%) relative to its moderate sample size (n ≈ 10,500). The substantial proportion of constrained features, combined with limited data, creates a scenario where constraints collectively restrict model flexibility more than the data can support. This pattern suggests that the ratio of constrained features to sample size, rather than absolute sample size alone, may be an important determinant of constraint cost – a finding with direct implications for practitioners deciding which features to constrain.

The variation across libraries, while modest, suggests some implementation-specific effects. CatBoost achieves a slightly lower average AUC PoM (0.6%) than XGBoost and LightGBM (0.9% each) and even attains negative PoM values – i.e., performance gains from constraints – on German Credit. These differences may reflect architectural variations in constraint enforcement: CatBoost's ordered boosting and symmetric trees can interact differently with monotonicity constraints than XGBoost's level-wise or LightGBM's leaf-wise growth. However, the much larger variation across datasets indicates that no library consistently dominates. Constraint costs are driven mainly by how well the imposed monotonic relationships align with the underlying data structure, with implementation details playing a secondary role.

Our wrong-sign diagnostic provides useful additional validation of the constraint specification. When constraints are applied with incorrect signs – for example, constraining income to increase default risk – the costs are substantial, ranging from 2.5% to 21.2% AUC PoM across datasets. This finding underscores that constraint costs depend critically on whether constraints align with true relationships in the data. When constraints are well-specified based on domain knowledge and validated through diagnostic tools like ICE plots, costs are generally small; when they are mis-specified, costs can be substantial. This validates our domain-knowledge-based approach to constraint specification and highlights the importance of careful constraint selection and validation.

Across datasets, monotonicity constraints impose consistently modest discrimination costs, while their impact on calibration is more variable. In our benchmark, AUC PoM typically lies between 0% and 2.9%, whereas Brier PoM ranges from essentially zero to double-digit increases. The Polish Bankruptcy dataset illustrates a case where constraints materially deteriorate both discrimination and calibration, suggesting that heavy constraint coverage in a complex, information-dense setting can restrict model flexibility in ways that harm both dimensions. By contrast, on larger and more parsimonious datasets, constraints often leave discrimination almost unchanged and have much smaller calibration effects. Taken together, these patterns indicate that the calibration–discrimination trade-off under monotonicity constraints is inherently context-dependent rather than governed by a universal rule.

These findings have direct implications for credit risk model development. For institutions with large datasets and clear domain knowledge for a subset of features, monotonicity constraints offer interpretability with little cost. In our benchmark, the two largest datasets ($n \gtrsim 100{,}000$) with moderate to high constraint coverage (up to about 60% of features) show AUC PoM well below 1%, typically around 0.1–0.2%. In similar settings, constraining a moderate set of well-understood features is unlikely to reduce AUC by more than about one percentage point, providing a practical benchmark for developers who balance regulatory interpretability with predictive performance.

For regulators and model validators, our results demonstrate that monotonicity constraints need not compromise predictive accuracy when appropriately specified. The substantial heterogeneity in constraint costs – and particularly the finding that constraints can be effectively “free” on large datasets – suggests that regulatory requirements for model interpretability need not come at the expense of discrimination performance. However, our results also highlight the importance of context-specific evaluation: blanket recommendations about constraint costs are inappropriate, and institutions should validate constraint effects on their specific datasets and feature sets.

Theoretically, our findings contribute to understanding when domain-knowledge constraints can help rather than hurt model performance. Our results are consistent with the view that monotonicity constraints act as a form of structural regularization: they reduce variance by ruling out economically implausible fits, while introducing only limited bias when constraints are aligned with the true data-generating process. In our large credit datasets, this manifests as an almost zero Price of Monotonicity – and occasionally slight gains – suggesting that, when economic priors are well-specified and constraint coverage is moderate, models can retain high predictive accuracy on unconstrained features while benefiting from constraint-induced variance reduction.

Our work also highlights important limitations and directions for future research. First, our analysis focuses on three popular gradient boosting libraries; extending comparisons to other implementations or alternative constraint-enforcing methods (e.g., monotonic neural networks, isotonic regression post-processing) would strengthen the evidence base. Second, our constraint specification relies on domain knowledge and train-only evidence; developing more systematic methods for constraint selection – perhaps based on statistical tests of monotonicity or cross-validation of constraint sets – could improve constraint specification in practice. Third, our analysis focuses on binary classification; extending to regression settings (e.g., loss given default, exposure at default) or multi-class problems would broaden applicability. Fourth, we examine feature-level monotonicity; exploring interaction constraints, partial monotonicity, or range-specific monotonicity could provide more flexible constraint specifications. Finally, understanding the relationship between constraint costs and downstream business metrics – capital requirements, portfolio risk, profitability – would strengthen the practical relevance of our findings.

Beyond credit risk, our methodological framework and empirical patterns may be informative for other high-stakes applications where interpretability and performance must be balanced. Healthcare decision support, criminal justice risk assessment, and environmental policy modeling all face similar tensions between model complexity and interpretability requirements. Our finding that, in credit risk, monotonicity constraints can be effectively "free" on large datasets when appropriately specified—and that their costs co-move with dataset characteristics and constraint coverage—suggests a template for evaluating analogous interpretability constraints in these domains.

In conclusion, this paper shows that, in several credit PD settings, appropriately specified monotonicity constraints on gradient boosting models can deliver interpretability with modest – and often negligible – performance costs. Across five datasets and three libraries, the Price of Monotonicity in AUC ranges from effectively zero to about 2.9%, with clear patterns in how these costs vary with dataset size, constraint coverage, library implementation, and the quality of constraint specification. For practitioners, the benchmarks provide empirical guidance on the magnitude of accuracy–interpretability trade-offs, helping to identify when constraints are likely to be acceptable versus when they become costly. For regulators, the results illustrate that, especially in large-scale portfolios with carefully chosen constraints, interpretability requirements need not entail large losses in predictive accuracy. For the broader interpretable machine learning community, the analysis clarifies when domain-knowledge constraints help or hurt model performance, with implications that extend beyond credit risk.

Rather than reflexively treating interpretability and performance as competing objectives, our results show that, at least in credit PD modeling, well-specified constraints can align model behavior with domain knowledge while preserving – and in some cases even enhancing – predictive accuracy. Recognizing this has important implications for how we develop, validate, and deploy machine learning models in high-stakes applications where both accuracy and interpretability are essential. As financial institutions increasingly deploy complex models for regulatory compliance and risk management, quantifying the empirical cost of interpretability constraints becomes a practical input into model governance and risk appetite, rather than merely an academic concern.

## Appendix A. Additional Tables

TABLE A.1 – COMPLETE LIST OF MONOTONICITY CONSTRAINTS BY DATASET AND FEATURE

| Dataset | Feature | Description | Constraint | Economic Reasoning |
|---|---|---|---|---|
| Taiwan Credit | X1 | Credit amount | Increasing (+1) | Larger credit limits/exposure correlate with higher absolute default risk |
| | X3 | Education level | Increasing (+1) | Lower education often associates with higher probability of default |
| | X4 | Marital status | Increasing (+1) | Single/other groups exhibit higher PD vs married on average |
| | X6 | Repayment status (month 1) | Increasing (+1) | More payment delays indicate worse payment history and higher default risk |
| | X7 | Repayment status (month 2) | Increasing (+1) | More payment delays indicate worse payment history and higher default risk |
| | X8 | Repayment status (month 3) | Increasing (+1) | More payment delays indicate worse payment history and higher default risk |
| | X9 | Repayment status (month 4) | Increasing (+1) | More payment delays indicate worse payment history and higher default risk |
| | X10 | Repayment status (month 5) | Increasing (+1) | More payment delays indicate worse payment history and higher default risk |
| | X11 | Repayment status (month 6) | Increasing (+1) | More payment delays indicate worse payment history and higher default risk |
| Polish Bankruptcy | A1 | Net profit / total assets | Decreasing (-1) | Higher profitability reduces bankruptcy risk |
| | A2 | Total liabilities / total assets | Increasing (+1) | Higher leverage increases bankruptcy risk |
| | A3 | Working capital / total assets | Decreasing (-1) | Higher working capital indicates better liquidity and lower risk |
| | A4 | Current assets / short-term liabilities | Decreasing (-1) | Higher ratio indicates better short-term liquidity and lower risk |
| | A5 | (Cash + securities + receivables - short-term liabilities) / (operating expenses - depreciation) × 365 | Decreasing (-1) | Higher ratio indicates better cash flow coverage and lower risk |
| | A6 | Retained earnings / total assets | Decreasing (-1) | Higher retained earnings indicate better financial stability and lower risk |
| | A7 | EBIT / total assets | Decreasing (-1) | Higher earnings indicate better profitability and lower risk |
| | A8 | Book value of equity / total liabilities | Decreasing (-1) | Higher equity relative to liabilities indicates better solvency and lower risk |
| | A9 | Sales / total assets | Decreasing (-1) | Higher asset turnover indicates better efficiency and lower risk |
| | A10 | Equity / total assets | Decreasing (-1) | Higher equity ratio indicates better financial stability and lower risk |
| | A12 | Gross profit / short-term liabilities | Decreasing (-1) | Higher profit relative to short-term debt indicates better ability to meet obligations |
| | A13 | (Gross profit + depreciation) / sales | Decreasing (-1) | Higher ratio indicates better profit margin and lower risk |
| | A14 | (Gross profit + interest) / total assets | Decreasing (-1) | Higher earnings relative to assets indicate better profitability and lower risk |
| | A16 | (Gross profit + depreciation) / total liabilities | Decreasing (-1) | Higher earnings relative to liabilities indicate better debt coverage and lower risk |
| | A17 | Total assets / total liabilities | Decreasing (-1) | Higher asset-to-liability ratio indicates better solvency and lower risk |
| | A18 | Gross profit / total assets | Decreasing (-1) | Higher gross profit margin indicates better profitability and lower risk |
| | A19 | Gross profit / sales | Decreasing (-1) | Higher gross margin indicates better operational efficiency and lower risk |
| | A22 | Profit on operating activities / total assets | Decreasing (-1) | Higher operating profit indicates better operational performance and lower risk |
| | A23 | Net profit / sales | Decreasing (-1) | Higher net profit margin indicates better profitability and lower risk |
| | A24 | Gross profit (3 years) / total assets | Decreasing (-1) | Higher cumulative profit indicates better long-term performance and lower risk |
| | A26 | (Net profit + depreciation) / total liabilities | Decreasing (-1) | Higher cash flow relative to liabilities indicates better debt service capacity and lower risk |
| | A27 | Profit on operating activities / financial expenses | Decreasing (-1) | Higher ratio indicates better ability to cover interest expenses and lower risk |
| | A29 | Logarithm of total assets | Decreasing (-1) | Larger firms (higher assets) tend to have lower bankruptcy risk |
| | A30 | (Total liabilities - cash) / sales | Increasing (+1) | Higher ratio indicates greater leverage relative to sales and higher risk |
| | A35 | Profit on sales / total assets | Decreasing (-1) | Higher profit indicates better profitability and lower risk |
| | A36 | Total sales / total assets | Decreasing (-1) | Higher asset turnover indicates better efficiency and lower risk |
| | A37 | (Current assets - inventories) / long-term liabilities | Decreasing (-1) | Higher ratio indicates better ability to cover long-term obligations and lower risk |
| | A39 | Profit on sales / sales | Decreasing (-1) | Higher profit margin indicates better profitability and lower risk |
| | A40 | (Current assets - inventory - receivables) / short-term liabilities | Decreasing (-1) | Higher ratio indicates better short-term liquidity and lower risk |
| | A42 | Profit on operating activities / sales | Decreasing (-1) | Higher operating margin indicates better operational performance and lower risk |

| | A46 | (Current assets - inventory) / short-term liabilities | Decreasing (-1) | Higher quick ratio indicates better short-term liquidity and lower risk |
|---|---|---|---|---|
| | A48 | EBITDA / total assets | Decreasing (-1) | Higher EBITDA indicates better operational cash flow and lower risk |
| | A49 | EBITDA / sales | Decreasing (-1) | Higher EBITDA margin indicates better operational efficiency and lower risk |
| | A50 | Current assets / total liabilities | Decreasing (-1) | Higher ratio indicates better ability to cover all liabilities and lower risk |
| | A51 | Short-term liabilities / total assets | Increasing (+1) | Higher ratio indicates greater short-term leverage and higher risk |
| | A55 | Working capital | Decreasing (-1) | Higher working capital indicates better liquidity and lower risk |
| | A56 | (Sales - cost of products sold) / sales | Decreasing (-1) | Higher gross margin indicates better operational efficiency and lower risk |
| | A58 | Total costs / total sales | Increasing (+1) | Higher cost ratio indicates lower efficiency and higher risk |
| | A59 | Long-term liabilities / equity | Increasing (+1) | Higher ratio indicates greater leverage and higher risk |
| | A60 | Sales / inventory | Decreasing (-1) | Higher inventory turnover indicates better efficiency and lower risk |
| | A64 | Sales / fixed assets | Decreasing (-1) | Higher fixed asset turnover indicates better efficiency and lower risk |
| German Credit (Statlog) | Attribute2 | Duration in months | Increasing (+1) | Longer loan duration increases default risk due to extended exposure |
| | Attribute5 | Credit amount | Increasing (+1) | Larger credit amounts increase default risk due to higher exposure |
| | Attribute8 | Installment rate (% of disposable income) | Increasing (+1) | Higher installment burden increases default risk by reducing financial flexibility |
| | Attribute11 | Present residence since | Decreasing (-1) | Longer residence indicates stability and lower risk of default |
| Give Me Some Credit | RevolvingUtilizationOfUnsecuredLines | Revolving utilization of unsecured lines | Increasing (+1) | Higher utilization indicates greater financial stress and higher default risk |
| | NumberOfTime30-59DaysPastDue | 30-59 days past due (not worse) | Increasing (+1) | More payment delays indicate higher default risk |
| | NumberOfTime60-89DaysPastDue | 60-89 days past due (not worse) | Increasing (+1) | More payment delays indicate higher default risk |
| | NumberOfTimes90DaysLate | 90+ days late | Increasing (+1) | More severe payment delays indicate higher default risk |
| | DebtRatio | Debt ratio | Increasing (+1) | Higher debt burden increases default risk by reducing financial flexibility |
| | MonthlyIncome | Monthly income | Decreasing (-1) | Higher income provides greater repayment capacity and lower default risk |
| Lending Club (Zenodo) | revenue | Annual income | Decreasing (-1) | Higher income provides greater repayment capacity and lower default risk |
| | fico_n | FICO credit score | Decreasing (-1) | Higher FICO score indicates better credit history and lower default risk |
| | dti_n | Debt-to-income ratio | Increasing (+1) | Higher debt burden increases financial stress and default risk |

*Notes*: This table provides the complete specification of all monotonicity constraints applied in this study. Constraints were determined based on economic reasoning, domain knowledge, prior credit risk literature, and train-only evidence (no test set leakage). Increasing constraints (+1) indicate that higher feature values correspond to higher default risk; decreasing constraints (-1) indicate that higher feature values correspond to lower default risk. Features with ambiguous or mixed evidence were left unconstrained.

TABLE A.2 – BEST HYPERPARAMETERS — XGBOOST

| Dataset | Model Type | Learning Rate | n_estimators | max_depth | reg_lambda |
|---|---|---|---|---|---|
| Taiwan Credit | Unconstrained | 0.01 | 500 | 5 | 5 |
| | Constrained | 0.01 | 500 | 6 | 10 |
| | Wrong-sign | 0.01 | 1,000 | 5 | 10 |
| Polish Bankruptcy | Unconstrained | 0.01 | 2,000 | 3 | 0 |
| | Constrained | 0.01 | 500 | 6 | 1 |
| | Wrong-sign | 0.05 | 500 | 3 | 5 |
| German Credit (Statlog) | Unconstrained | 0.01 | 500 | 3 | 5 |
| | Constrained | 0.01 | 500 | 5 | 10 |
| | Wrong-sign | 0.01 | 500 | 3 | 10 |
| Give Me Some Credit | Unconstrained | 0.01 | 2,000 | 3 | 10 |
| | Constrained | 0.01 | 500 | 9 | 10 |
| | Wrong-sign | 0.03 | 200 | 5 | 10 |
| Lending Club (Zenodo) | Unconstrained | 0.01 | 1,000 | 5 | 5 |
| | Constrained | 0.01 | 1,000 | 5 | 5 |
| | Wrong-sign | 0.03 | 1,000 | 3 | 5 |

TABLE A.3 – BEST HYPERPARAMETERS — LIGHTGBM

| Dataset | Model Type | Learning Rate | n_estimators | num_leaves | min_child_samples |
|---|---|---|---|---|---|
| Taiwan Credit | Unconstrained | 0.01 | 500 | 31 | 30 |
| | Constrained | 0.01 | 500 | 31 | 50 |
| Polish Bankruptcy | Unconstrained | 0.01 | 500 | 63 | 100 |
| | Constrained | 0.03 | 200 | 31 | 30 |
| German Credit (Statlog) | Unconstrained | 0.01 | 1,000 | 31 | 100 |
| | Constrained | 0.05 | 200 | 31 | 100 |
| Give Me Some Credit | Unconstrained | 0.01 | 500 | 31 | 100 |
| | Constrained | 0.01 | 500 | 31 | 100 |
| Lending Club (Zenodo) | Unconstrained | 0.01 | 1,000 | 31 | 10 |
| | Constrained | 0.01 | 1,000 | 31 | 50 |

TABLE A.4 – BEST HYPERPARAMETERS — CATBOOST

| Dataset | Model Type | Learning Rate | n_estimators | depth | l2_leaf_reg |
|---|---|---|---|---|---|
| Taiwan Credit | Unconstrained | 0.01 | 1,000 | 6 | 10 |
| | Constrained | 0.01 | 500 | 8 | 20 |
| Polish Bankruptcy | Unconstrained | 0.03 | 2,000 | 5 | 20 |
| | Constrained | 0.05 | 1,500 | 5 | 20 |
| German Credit (Statlog) | Unconstrained | 0.01 | 500 | 8 | 10 |
| | Constrained | 0.01 | 500 | 9 | 5 |
| Give Me Some Credit | Unconstrained | 0.01 | 1,000 | 9 | 20 |
| | Constrained | 0.1 | 200 | 6 | 20 |
| Lending Club (Zenodo) | Unconstrained | 0.05 | 1,500 | 6 | 10 |
| | Constrained | 0.1 | 1,000 | 6 | 5 |

*Notes*: This table reports the best hyperparameters selected via 3-fold cross-validation grid search on the training set, scoring by negative LogLoss. The same grid search space was used for unconstrained and constrained models to ensure fair comparison. Hyperparameters were selected independently for each dataset, library, and model type. Wrong-sign models (XGBoost only) were tuned separately as a diagnostic.

# Appendix B. Additional Figures

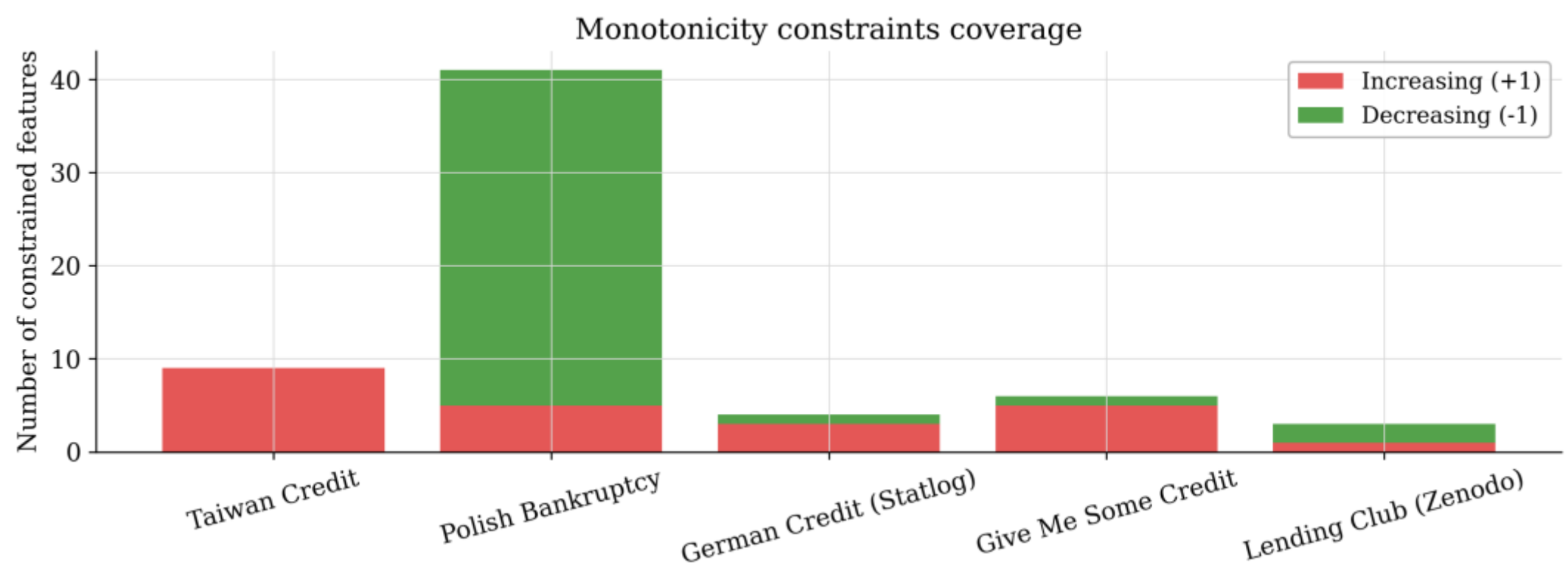

FIGURE B.1 – DISTRIBUTION OF MONOTONICITY CONSTRAINTS BY DATASET AND DIRECTION

*Notes*: Stacked bar chart showing the number of features with monotonicity constraints by dataset and constraint direction. Red segments (bottom) represent increasing constraints (+1): higher feature value → higher default risk. Green segments (top) represent decreasing constraints (-1): higher feature value → lower default risk. Total bar height indicates total constrained features per dataset. Polish Bankruptcy has the highest coverage (41 features, predominantly decreasing), reflecting financial ratio constraints. Taiwan Credit has 9 features (all increasing), consistent with payment history constraints. Coverage varies from 3 (Lending Club) to 41 (Polish Bankruptcy) features, reflecting dataset characteristics and selective constraint application based on economic reasoning.

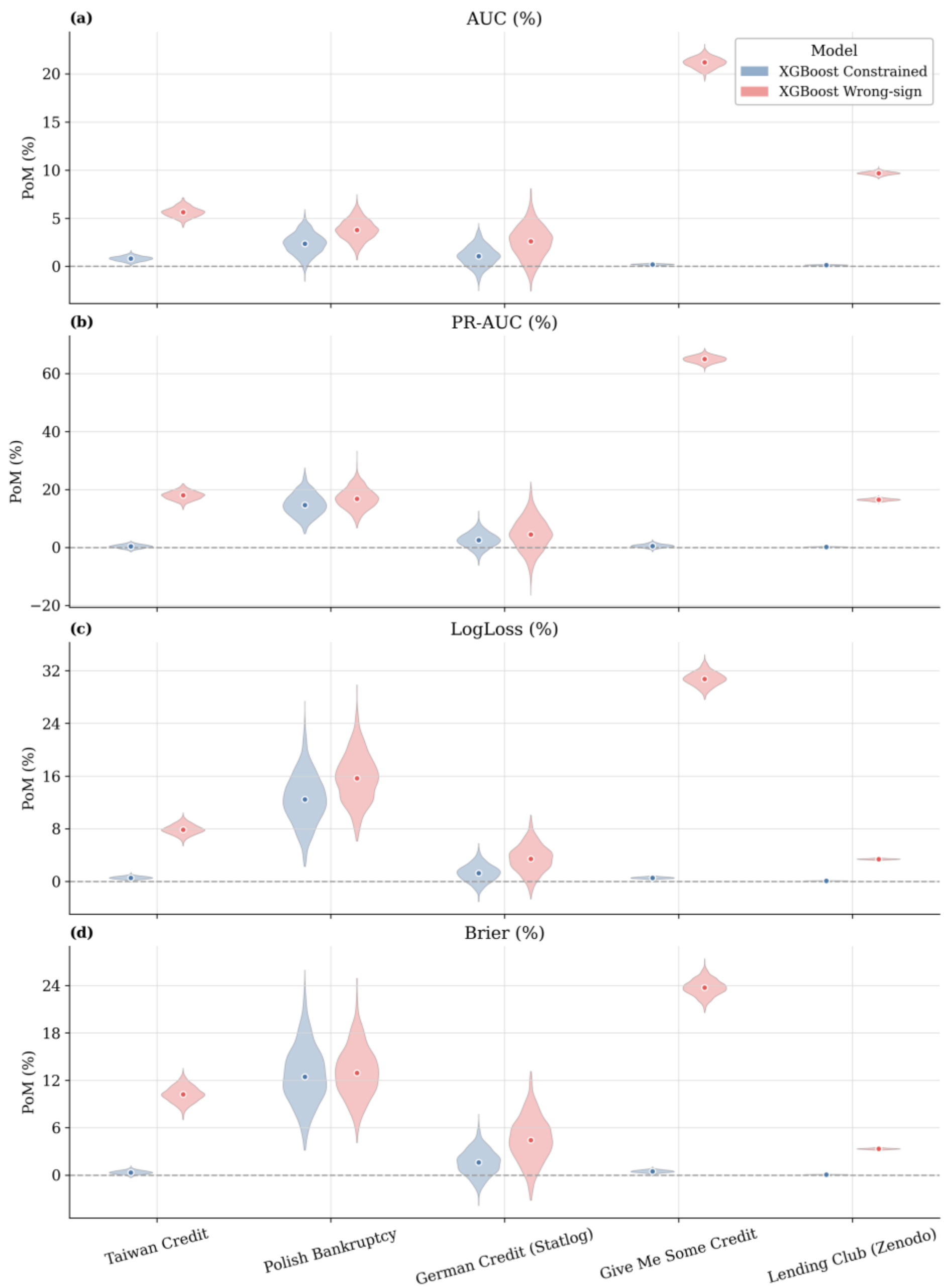

FIGURE B.2 – CORRECTLY SPECIFIED VS. MIS-SPECIFIED CONSTRAINTS

*Notes*: Multi-panel violin plots comparing Price of Monotonicity (PoM) distributions for XGBoost models with correctly specified monotonicity constraints (blue) versus intentionally mis-specified wrong-sign constraints (red) across five credit risk datasets. Panels show (a) AUC, (b) PR-AUC, (c) LogLoss, and (d) Brier score. PoM is calculated relative to unconstrained XGBoost performance. Distributions are based on 1,000 paired bootstrap replicates. The dashed grey line at 0% indicates no performance difference.